\documentclass{article}

\usepackage[preprint]{neurips_2025}

\usepackage[utf8]{inputenc} 
\usepackage[T1]{fontenc}    
\usepackage{hyperref}       

\usepackage{url}            
\usepackage{booktabs}       
\usepackage{amsfonts}       
\usepackage{nicefrac}       
\usepackage{microtype}      
\usepackage{xcolor}         

\usepackage{amsmath}
\usepackage{pifont}
\usepackage{textcomp}
\usepackage{array}
\usepackage{multirow}
\usepackage{graphicx}
\usepackage{amssymb}

\usepackage{fontawesome5}  
\usepackage{hyperref}      
\title{OJBench: A Competition Level Code Benchmark For Large Language Models}

\author{
Zhexu Wang$^1$\thanks{Equal contribution.} , 
Yiping Liu$^2$\footnotemark[1] , 
Yejie Wang$^1$, 
Wenyang He$^3$, 
Bofei Gao$^4$, 
Muxi Diao$^1$,\\ 
{\bf Yanxu Chen$^1$}, 
{\bf Kelin Fu$^4$}, 
{\bf Flood Sung$^5$}, 
{\bf Zhilin Yang$^5$}, 
{\bf Tianyu Liu$^4$}\thanks{Project lead.}, 
{\bf Weiran Xu$^1$}\thanks{Corresponding author.} \\
$^1$Beijing University of Posts and Telecommunications  $^2$Tsinghua University \\
$^3$University of Chinese Academy of Sciences $^4$Peking University $^5$Moonshot AI \\
\texttt{\{wangzhexu, xuweiran\}@bupt.edu.cn}\\
\href{https://github.com/He-Ren/OJBench}{\faGithub\ Code} \quad
\href{https://he-ren.github.io/OJBench-Leaderboard/}{\textcolor{orange}{\faTrophy}\ Leaderboard}\thanks{We will continuously update the performance of newly released models on our leaderboard.} \\
}

\begin{document}

\maketitle

\begin{abstract}
    Recent advancements in large language models (LLMs) have demonstrated significant progress in math and code reasoning capabilities. However, existing code benchmark are limited in their ability to evaluate the full spectrum of these capabilities, particularly at the competitive level. To bridge this gap, we introduce OJBench, a novel and challenging benchmark designed to assess the competitive-level code reasoning abilities of LLMs. OJBench comprises 232 programming competition problems from NOI and ICPC, providing a more rigorous test of models' reasoning skills. We conducted a comprehensive evaluation using OJBench on 37 models, including both closed-source and open-source models, reasoning-oriented and non-reasoning-oriented models. Our results indicate that even state-of-the-art reasoning-oriented models, such as o4-mini and Gemini-2.5-pro-exp, struggle with highly challenging competition-level problems. This highlights the significant challenges that models face in competitive-level code reasoning.
\end{abstract}

\section{Introduction}
With the emergence of long chain-of-thought (CoT) models such as OpenAI-o1, o4-mini\cite{jaech2024openai}, Gemini-2.5-pro-exp\cite{GoogleDe55:online}, DeepSeek-R1\cite{guo2025deepseek} and Qwen3\cite{qwen3}, large language models oriented towards reasoning have demonstrated advanced reasoning capabilities\cite{cobbe2021training}, mathematics\cite{hendrycks2021measuring}, physics \cite{welbl2017crowdsourcing}, and formal proof \cite{kimina_prover_2025} \cite{deepseek9:online}, but  also achieved significant progress in areas like code generation and software engineering\cite{fan2023large}. Math and code are two domains commonly used to evaluate the reasoning abilities of models. Although the field of mathematics has numerous competition-level benchmarks, such as Omni-Math \cite{gao2024omni}, AIME\cite{Pagenotf85:online}, and LiveAoPSBench\cite{mahdaviaops}, the field of code still lacks competition-level benchmarks that can pose a challenge to models. Currently, researchers primarily evaluate models' performance on complex algorithmic programming tasks using LiveCodeBench\cite{jain2024livecodebench}. However, most problems in LiveCodeBench are of interview difficulty level and focus on basic algorithms. Many models have achieved similar high performance on LiveCodeBench, highlighting the need for more challenging benchmarks to better differentiate their capabilities and uncover limitations, thereby guiding the advancement of future code LLMs.

An ideal scenario is to use real competitive programming tasks, which are often designed for top-tier programmers selected globally. Inspired by the practice of testing models' coding abilities on the CodeForces platform by models such as OpenAI-o1\cite{jaech2024openai}, o3\cite{Day12ofS75:online}, and DeepSeek-R1\cite{guo2025deepseek}, the recent study CodeElo\cite{quan2025codeelo} has explored how to evaluate the competitive coding abilities of different models.  However, this approach relies on simulated submissions on the CodeForces platform, where the difficulty of problems varies widely. This variability makes it difficult to effectively challenge the model's competition-level coding capabilities.

To bridge this gap, we propose OJBench, a competition-level code benchmark comprising 232 problems that can effectively assess models' code reasoning abilities in competitive programming tasks. Using OJBench, we evaluated 37 models, including the Coder models trained on large-scale code corpora and reasoning-oriented models trained with reinforcement learning on a wide range of reasoning tasks. We found that reasoning-oriented models significantly outperformed non-reasoning-oriented models in competitive coding tasks, and open-source models still lag behind closed-source models in terms of code reasoning ability. Additionally, we found that for most long-CoT models, using CPP as the programming language performs better than Python. Moreover, these models can leverage feedback from the code execution environment to refine their error solutions, thereby enhancing their overall performance.

In summary, our main contributions are as follows:

1) We introduce OJBench, a competition-level code benchmark that encompasses 232 top-tier algorithmic competition problems from human contests.

2) Utilizing OJBench, we evaluated 37 models and revealed the limitations of current models in complex code reasoning tasks.

3) We investigated the performance of models using different programming languages in complex programming tasks and experimented with models correcting their erroneous solutions based on environmental feedback, thereby providing guidance for future code LLM development.

\section{OJBench}
In Section \ref{subsection:2.1}, we introduced the sources of the problems in OJBench and the data collection process. In Section \ref{subsection:2.2}, we demonstrated how we assigned difficulty labels to each problem. In Sections \ref{subsection:2.3} and \ref{subsection:2.4}, we respectively elaborated on the evaluation methods employed by OJBench and its support for dual-language assessment in both Python and CPP.

\subsection{Data Source and Data Collection}
\label{subsection:2.1}
\begin{figure*}[t]
    \centering
    \resizebox{\linewidth}{!}{
    \includegraphics{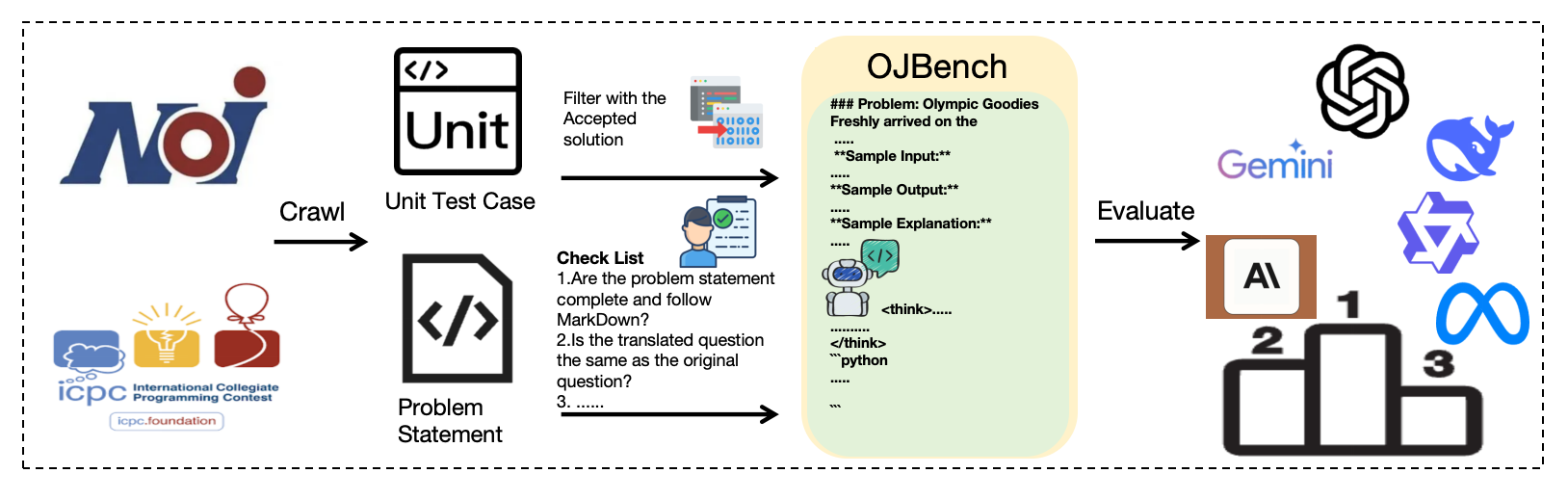}}

    \caption{The overall data collection and filtering process of OJBench.}
    \label{fig:main}
    \vspace{-0.3cm}
\end{figure*}

The overall data collection and data filtering process of OJBench is shown in Figure\ref{fig:main}.

\textbf{Data Source. }  First we reviewed existing benchmarks, such as USACO\cite{shi2024can}, contain problems exclusively from the USA Computing Olympiad. LiveCodeBench\cite{jain2024livecodebench} and codeELO\cite{quan2025codeelo}, on the other hand, utilize problems from the LeetCode and Codeforces platforms respectively. To minimize overlap with existing benchmarks and incorporate more challenging competitive programming problems into our dataset, we selected problems from China's National Olympiad in Informatics (NOI) and International Collegiate Programming Contest (ICPC) as our data sources. For detailed information on NOI and ICPC, please refer to Appendix \ref{appendix:Details of NOI and ICPC}.

\textbf{Data Collection. }  We collected NOI and ICPC questions from the Logu competition platform and the ICPC official website respectively. Each data sample includes problem descriptions in Markdown format and the comprehensive test cases released by the competition organizers. These extensive test cases provides a solid foundation for our evaluation set, thereby ensuring its validity.
Subsequently, we conducted a simple data filtering process:

1) We validated the collected test cases using the correct code submissions from contestants, filtering out data samples that failed to pass the judge. This operation effectively avoided issues related to incomplete or erroneous test cases.

2) In the initial dataset, not all problems could be judged by directly comparing the output with the correct answer, as some problems have non-unique correct outputs and require specific judging programs to assess the correctness of the answers. However, for certain problems, these special judging programs are often unavailable and difficult to write. Therefore, we filtered out the problems that required a special judge in the original dataset.

Finally, we used GPT-4o to translate the problem descriptions of NOI into English, followed by manual verification to ensure that the translated problems retained their original meaning and maintained the correct format. The statistics of OJBench are presented in Table  \ref{tab:difficulty}.

\begin{table}[t]
    \centering
    \resizebox{0.75\textwidth}{!}{
        \begin{tabular}{c|c|c|c|c|c}
        \toprule
        Source & Count & Easy & Mediun & Hard  &  Average Tests \\
        \midrule
        NOI & 159 & 20 & 53 & 86 & 17.21 \\
        ICPC & 73 & 16 & 26 & 31 & 63.60 \\
        \midrule
        Total & 232 & 36 & 79 & 117 & 31.81 \\
        \bottomrule
        \end{tabular}
    }
    \vspace{10pt}
    \caption{The statistics of problems collected in OJBench.}
    \vspace{-15pt}
    \label{tab:difficulty}
\end{table}

\subsection{Difficulty Classification}
\label{subsection:2.2}

To better evaluate the performance of models across problems of varying difficulty levels, we categorized the problems in OJBench into three distinct difficulty levels: Easy, Medium, and Hard.

\textbf{NOI. }   For problems sourced from NOI, we utilized the difficulty ratings obtained from the competition platform for categorization. Specifically, the difficulty of each problem is annotated through voting by contestants who successfully solved the problem, which can effectively reflect the true difficulty level of the problem. The difficulty of each problem is quantified on a scale from 0 to 7, with higher values indicating greater difficulty. The difficulty range of OJBench problems spans from 2 to 7. We categorized problems with a difficulty level of 2-3 as Easy, 4-5 as Medium, and 6–7 as Hard.

\textbf{ICPC. }   For problems sourced from ICPC,due to the lack of an official difficulty classification, we crawled the ranking information from real competitions. For each problem, we considered four metrics: total submission count (total\_submission), total number of accepted solutions (total\_passed), total number of participating teams (total\_team), and the number of teams that attempted the problem (attempted\_team). We calculated the difficulty score for each problem using the following formula:

\begin{equation*}
\text{score} = \left( \frac{\text{total\_passed}}{\text{total\_submission}} \right) \times \left( \frac{\text{attempted\_team}}{\text{total\_team}} \right)
\end{equation*}

The ratio 
\( \frac{\text{total\_passed}}{\text{total\_submission}} \)
measures the success rate of teams attempting to solve the problem. In real programming contests, teams often assess the difficulty of a problem before deciding whether to attempt it. Therefore, the ratio 
\(  \frac{\text{attempted\_team}}{\text{total\_team}} \)
indicates the perceived difficulty level of the problem by the participating teams. This score effectively captures the true difficulty of the problem. Specifically, we classified problems with a score of 0.4 or higher as Easy, those with a score of 0.1 or lower as Hard, and the rest problems as Medium.

\subsection{Evaluation method}
\label{subsection:2.3}
\begin{figure*}[t]
    \centering
    \vspace{-5pt}
    \resizebox{0.70\linewidth}{!}{
    \includegraphics{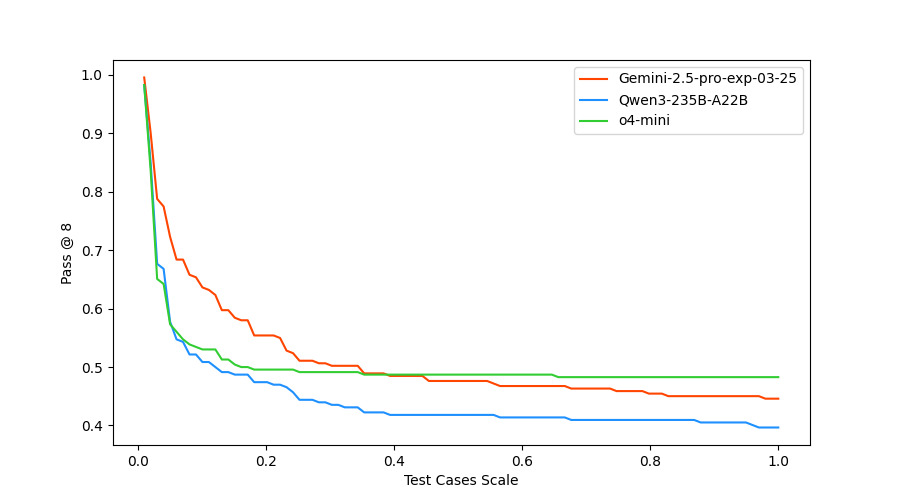}}
    \vspace{-0.3cm}
    \caption{Performance decline of models with increasing test case scale on OJBench.}
    \label{fig:test_case_scale}
\end{figure*}

\textbf{Judge Base on the test case. }  In programming competitions, test cases are commonly used to verify the correctness of code. In programming tasks, models are provided with problem descriptions and input-output examples, and are required to generate correct solutions. The correctness of these solutions is evaluated based on a set of test cases (input-output pairs). Previous research has indicated that using a limited number of test cases for evaluation can easily lead to false positives results\cite{yang2025probench}. Similarly, in our experiments, we observed that using a small number of test cases for validation tends to produce false-positive results. As shown in Figure \ref{fig:test_case_scale}, the performance of all models decreases with an increasing number of test cases. Even powerful reasoning-oriented  models, such as o4-mini\cite{Introduc30:online}, Gemini-2.5-pro-exp-03-25 \cite{GeminiGo69:online}, Qwen3-235B-A22B\cite{qwen3}, exhibit this trend. To rigorously validate the robustness of the code, we assess the correctness of the programs generated by the models on the entire set of test cases. A solution is considered correct only if it passes all test cases.

\vspace{0.2cm}
\subsection{Dual Programming Language Assessment.}
 \label{subsection:2.4}

In the existing competition-level code benchmarks, such as LiveCodeBench \cite{jain2024livecodebench} and APPS\cite{hendrycks2021measuring}, the majority of evaluations are conducted solely on Python. Although Python is the most commonly used programming language for most models, in competitive programming, human contestants predominantly utilize CPP \cite{quan2025codeelo}. In fact, for top-tier competitive programming problems, which have strict algorithmic time complexity requirements, CPP is the preferred language. Evaluating models solely on Python fails to accurately assess their performance on such problems. Therefore, OJBench supports evaluation in both Python and CPP programming languages. This dual-language support enables us to deeply investigate the performance differences when models use Python or CPP, thereby offering guidance for the development of future code reasoning-oriented models.Our findings reveal that for most reasoning-oriented models, using CPP as the programming language achieves better performance compared to Python. This result is consistent with human experience, because under the same conditions, CPP runs more efficiently than Python, making it more appropriate to use CPP to solve competition-level code problems with strict algorithmic time complexity requirements.  For detailed experimental results and analysis, please refer to Section \ref{subsection:Comparison between CPP and Python}.

\section{Evaluation on Existing LLMs}
\subsection{Experimental Setup}
\textbf{Metrics. }  We employ \textbf{Pass@n} \cite{kulal2019spoc}\cite{chen2021evaluating}as our evaluation metric. Which means that if a model gives n answers to the same problem and at least one of the answers is correct, the model is considered to have passed the problem. Specifically, we use API or VLLM\cite{kwon2023efficient} to generate eight candidate solutions for each problem. During evaluation, we assess the models using the hyper parameters officially recommended by each model, including temperature, top\_k, and top\_p. For reasoning-oriented models, which often generate long chains of thought, we set the maximum number of tokens (max\_tokens) to 64k. For non-reasoning-oriented models, we set max\_tokens to the default context length of the model.

\vspace{0.3cm}
\textbf{Models.} Using OJBench, we conducted a comprehensive evaluation of a wide range of models. For non-reasoning-oriented open-source models, we selected Qwen2.5-Coder-Instruct-7B/14B/32B\cite{hui2024qwen2}, DeepSeek-Coder-Instruct-V2/Lite\cite{zhu2024deepseek}, DeepSeek-Code-Instruct-6.7B/33B\cite{guo2024deepseek}, DeepSeek-V3-0324\cite{liu2024deepseek}, and CodeLlama-Instruct-7B/13B/34B/70B\cite{grattafiori2024llama}. For reasoning-oriented open-source models, we evaluated DeepSeek-R1\cite{guo2025deepseek}, QWQ-32B\cite{qwq32b}, Qwen3-32B\cite{qwen3}, Qwen3-30B-A3B, Qwen3-235B-A22B, DeepSeek-R1-Distill-Qwen-1.5B/7B/14B/32B, DeepSeek-R1-Distill-Llama-8B/70B, and Olympic-Coder-7B/32B\cite{openr1Ol12:online}. Among closed-source models, we assessed Claude-3.5-sonnet-20241022\cite{teamintroducing}, Claude-3.7-sonnet-20250219, Claude-3.7-sonnet-20250219-Thinking, GPT3.5-Turbo, GPT4-Turbo, GPT4o-2024-11-20\cite{hurst2024gpt}, o1-mini\cite{jaech2024openai}, o1-2024-12-17, o3-mini\cite{Day12ofS75:online}, o4-mini, Gemini-2.0-Flash-Thinking-exp\cite{GoogleDe55:online}, and Gemini-2.5-pro-exp-03-25\cite{GeminiGo69:online}.

\begin{table}[t]
    \centering
    \resizebox{\textwidth}{!}{
        \begin{tabular}{c|cccc|ccc}
        \toprule
        Model & \multicolumn{4}{c|}{Pass@} & \multicolumn{3}{c}{Pass Rate @} \\ \cmidrule{2-8}
        & 1@Py & 8@Py & 1@CPP & 8@CPP & Easy & Mid & Hard \\ 
        
        \midrule
        \multicolumn{8}{c}{\textit{Non-Reasoning-oriented Open-source LLM}} \\
        \midrule

        Qwen2.5-Coder-7B & 3.50 & 4.74 & 2.64 & 4.74 & 20.49 & 0.95 & 0.00\\
        Qwen2.5-Coder-14B & 6.30 & 10.34 & 4.53 & 8.62 & 35.07 & 2.53 & 0.00 \\
        Qwen2.5-Coder-32B & 5.77 & 9.05 & 6.36 & 12.93 & 30.56 & 3.01 & 0.00\\
        
        CodeLlama-7B & 0.00 & 0.00 & 0.00 & 0.00 & 0.00 & 0.00 & 0.00\\
        CodeLlama-13B & 0.59 & 1.29 & 0.86 & 1.29 & 3.82 & 0.00 & 0.00 \\
        CodeLlama-34B & 0.11 & 0.86 & 0.48 & 0.86 & 0.69 & 0.00 & 0.00\\
        CodeLlama-70B & 2.53 & 4.74 & 0.92 & 3.02 & 13.54 & 1.27 & 0.00 \\

        DeepSeek-Coder-6.7B  &1.99 & 3.88 & 1.19 & 3.45 & 10.07 & 1.27 & 0.00 \\
        DeepSeek-Coder-33B  &2.64 & 6.47 & 2.59 & 6.47 & 14.24 & 1.27 & 0.00\\
        DeepSeek-Coder-V2-Lite  & 4.80 & 7.76 & 4.04 & 7.76 & 26.04 & 2.22 & 0.00\\
        DeepSeek-Coder-V2  & 8.24 & 11.21 & 8.67 & 14.22 & 44.79 & 3.80 & 0.00 \\
        DeepSeek-V3-0324 & \textcolor{blue}{25.54} & \textcolor{blue}{32.33} & \textcolor{blue}{22.95} & \textcolor{blue}{34.05} & \textcolor{blue}{78.47} & \textcolor{blue}{33.70} & \textcolor{blue}{3.74} \\
        
        \midrule
        \multicolumn{8}{c}{\textit{Reasoning-oriented Open-source LLM}} \\ 
        \midrule

        QWQ-32B & 19.02 & 30.17 & 19.77 & 31.90 & 69.10 & 20.89 & 2.35 \\
        Qwen3-32B & 14.92 & 32.76 & 15.09 & 31.90 & 53.12 & 17.25 & 1.60\\
        Qwen3-30B-A3B & 15.84 & 29.31 & 12.98 & 25.00 & 55.21 & 19.78 & 1.07 \\
        Qwen3-235B-A22B & 25.97 & \textcolor{green}{40.52} & \textcolor{green}{26.08} & \textcolor{green}{39.22} & 76.39 & 35.13 & \textcolor{green}{4.27} \\
        Olympic-Coder-7B & 8.78 & 18.10 & 8.24 & 18.97 & 42.36 & 6.17 & 0.21\\
        Olympic-Coder-32B & 15.41 & 29.31 & 16.27 & 31.03 & 54.51 & 18.67 & 1.18 \\
        DeepSeek-R1-Distill-Qwen-1.5B &2.59 & 6.47 & 0.05 & 0.43 & 13.54 & 1.42 & 0.00 \\
        DeepSeek-R1-Distill-Qwen-7B &8.94 & 15.09 & 0.97 & 2.59 & 39.93 & 8.07 & 0.00 \\
        DeepSeek-R1-Distill-Qwen-14B & 14.17 & 23.28 & 5.87 & 13.79 & 56.94 & 14.40 & 0.85 \\
        DeepSeek-R1-Distill-Qwen-32B &17.83 & 30.17 & 10.40 & 24.14 & 62.50 & 19.94 & 2.67 \\
        DeepSeek-R1-Distill-Llama-8B & 8.46 & 16.38 & 1.72 & 6.47 & 40.28 & 5.54 & 0.64\\
        DeepSeek-R1-Distill-Llama-70B & 16.38 & 28.02 & 10.02 & 23.71 & 61.81 & 18.20 & 1.18\\
        DeepSeek-R1 & \textcolor{green}{26.02} & 37.07 & 25.97 & 38.36 & \textcolor{green}{78.47} & \textcolor{green}{35.44} & 3.53 \\

        \midrule
        \multicolumn{8}{c}{\textit{Closed-source LLM}} \\ 
        \midrule
        
        Claude-3.5-sonnet-20241022 & 10.40 & 17.67 & 13.09 & 23.28 & 47.57 & 8.54 & 0.21 \\
        Claude-3.7-sonnet-20250219 &4.24 & 7.76 & 15.41 & 23.71 & 19.79 & 1.62 & 0.93  \\
        Claude-3.7-sonnet-20250219-Thinking &18.27 & 25.00 & 14.71 & 23.28 & 68.75 & 20.41 & 1.28 \\
        GPT3.5-Turbo & 5.60 & 9.05 & 4.26 & 8.62 & 31.25 & 1.74 & 0.32 \\
        GPT4-Turbo &9.97 & 16.81 & 9.70 & 18.53 & 46.18 & 7.28 & 0.64 \\
        GPT-4o-20241120 & 10.02 & 15.95 & 10.34 & 16.81 & 50.69 & 5.85 & 0.32 \\
        o1-mini & 21.61 & 33.62 & 25.97 & 37.50 & 68.75 & 30.22 & 1.28 \\
        o1-20241217& 26.45 & 35.78 & 33.24 & 47.84 & 81.94 & 35.92 & 2.99 \\
        o3-mini & 31.79 & 46.12 & 40.25 & 52.16 & \textcolor{red}{84.03} & 44.94 & 6.84 \\
        o4-mini & 33.30 & \textcolor{red}{48.71} & \textcolor{red}{46.12} & \textcolor{red}{61.21} & 83.33 & 51.27 & 5.77 \\
        Gemini-2.0-Flash-Thinking-exp &13.58 & 22.41 & 11.80 & 20.26 & 54.86 & 13.45 & 0.96 \\
        Gemini-2.5-pro-exp-03-25  & \textcolor{red}{38.91} & \textcolor{red}{48.71} & 44.26 & 56.47 & 83.68 & \textcolor{red}{61.71} & \textcolor{red}{9.48}\\   
        \bottomrule
        \end{tabular}
    }
    \vspace{10pt}
    \caption{The main results of different models on OJBench using Python(Py) and CPP programming.Pass@1@Py and Pass@8@Py represent Pass@1 and Pass@8 in Python respectively, and the same is true for CPP. The highest scores of open source non-reasoning-oriented models, open source reasoning-oriented models, and closed source models are marked in \textcolor{blue}{blue}, \textcolor{green}{green}, and \textcolor{red}{red} respectively.}
    \vspace{-20pt}
    \label{tab:Main results}
\end{table}


\subsection{Main results}
Table \ref{tab:Main results} presents the overall performance of all models on OJBench, and the pass@1 scores of the models using Python at different difficulty levels.

\textbf{Overall Performance. }  In general, closed-source models outperform open-source models. Among open-source reasoning-oriented  models, Qwen3-235B-A22B and DeepSeek-R1 demonstrate the best performance, surpassing o1-mini and approaching the performance of o1-20241217. In the closed-source category, previous models such as GPT3.5 and GPT4 showed limitations in competition-level code reasoning tasks. However, models that underwent large-scale post-training specifically for reasoning tasks, such as o4-mini and Gemini-2.5-pro-exp-03-25, exhibit state-of-the-art performance. For models within the same family but of different sizes, performance improves with an increase in parameter count. Overall, models designed for reasoning tasks consistently outperform non-reasoning-oriented models, highlighting the significant advantages of post-training methods like reinforcement learning and distillation in enhancing reasoning capabilities.

\textbf{Pass@n. }  As the number of samples n increases, the pass@n of all models consistently rises. Among open-source models, Qwen3-235B-A22B and DeepSeek-R1 exhibit a pass@8 improvement of 14.55 and 11.05 in python, respectively, compared to pass@1, surpassing the performance of o1-20241217 in the pass@8. For closed-source models, pass@8 also shows significant improvement over pass@1. For instance, o3-mini, o4-mini, and Gemini-2.5-pro-exp-03-25 achieve absolute improvements of 14.33, 15.41, and 9.8, respectively. This indicates that models are capable of exploring diverse approaches to problem solving.

\textbf{Pass Rate at Different Difficulty Levels.}  Table \ref{tab:Main results} also  presents the pass rates of the models implemented in Python across problems of varying difficulty levels in a single response. It is evident that even problems of simple difficulty pose a challenge for the majority of models. For problems at the Hard difficulty level, the pass rates of almost all non-reasoning-oriented models are zero. Notably, even the DeepSeek-V3-0324 model with an exceptionally large parameter scale of 671B achieved a pass rate of only 3.74\%. In contrast, most reasoning-oriented models perform well on problems of easy difficulty. This indicates that merely relying on large-scale pretraining is insufficient to enhance models' performance on competition-level coding problems. Instead, reinforcement learning and distillation from powerful reasoning-oriented  models hold significant potential for improving models' code reasoning capabilities.
For most reasoning-oriented models, performance on medium and hard problems is relatively lower but provides better differentiation. Hard problems effectively distinguish the top reasoning-oriented  models, such as o4-mini, Gemini-2.5-pro-exp-03-25, o3-mini, Qwen3-235B-A22B, and DeepSeek-R1. These results indicate that OJBench can effectively evaluate models' code reasoning abilities and is suitable for assessing the capabilities of future code LLMs.

\section{Analysis}
\subsection{Analysis of the overall difficulty of OJBench}
To objectively evaluate the overall difficulty level of OJBench, we conducted a comparative analysis by benchmarking it against other existing benchmarks. The performance of state-of-the-art models on these benchmarks was used as the basis for comparison. To the best of our knowledge, the two most recent competition-level benchmarks are CodeElo and Probench. However, the publicly available evaluation results from these benchmarks do not cover the latest reasoning-oriented models, such as o4-mini and Gemini-2.5-pro-exp-03-25. Therefore, we selected the widely used LiveCodeBench as the reference benchmark for comparison. Specifically, we directly utilized the evaluation results of o4-mini (low), Gemini-2.5-pro, and Qwen3-235B-A22B published on the LiveCodeBench leaderboard. These results were derived from a subset of data between January 1, 2025, and May 1, 2025.

The comparison results are shown in Table \ref{tab:comparison on ojbench and lcb}. Although these three models have excellent performance on LiveCodeBench, their performance on OJBench is significantly lower than LiveCodeBench, which shows that the overall difficulty of our proposed OJBench is higher. However, it is worth noting that we can only roughly estimate the difficulty of OJBench relative to LiveCodeBench based on the performance of these models, and cannot determine the magnitude of the difficulty difference. Meanwhile, we encourage researchers to include more benchmarks in the evaluation of models to provide more accurate and comprehensive evaluation results.
\begin{table}[t]
    \centering
    \resizebox{0.85\textwidth}{!}{
        \begin{tabular}{c|c|c|c|c|c}
        \toprule
        Model & Benchmark & Pass@1 & Easy-Pass@1 & Mediun-Pass@1 & Hard-Pass@1  \\
        \midrule
        o4-mini(low) & LCB & 63.70 & 97.80 & 76.40 & 36.60  \\
        o4-mini     & OJBench & 33.30 & 83.33 & 51.27 & 5.77  \\
        \midrule
        Gemini-2.5-pro &  LCB & 65.90 & 100 & 74.50 & 41.50  \\
        Gemini-2.5-pro-exp-03-25 & OJBench & 38.91 & 83.68 & 61.71 & 9.48  \\
        \midrule
        Qwen3-235B-A22B &  LCB & 56.60 & 100 & 72.70 & 28.00  \\
        Qwen3-235B-A22B & OJBench & 25.97 & 76.39 & 35.13 & 1.07  \\
        \bottomrule
        \end{tabular}
    }
    \vspace{10pt}
    \caption{Comparison of OJBench and LiveCodeBench based on model performance.}
    \vspace{-10pt}
    \label{tab:comparison on ojbench and lcb}
\end{table}

\subsection{Comparison between CPP and Python}
\label{subsection:Comparison between CPP and Python}
\begin{figure*}[h]
    \centering
    \resizebox{\linewidth}{!}{
    \includegraphics{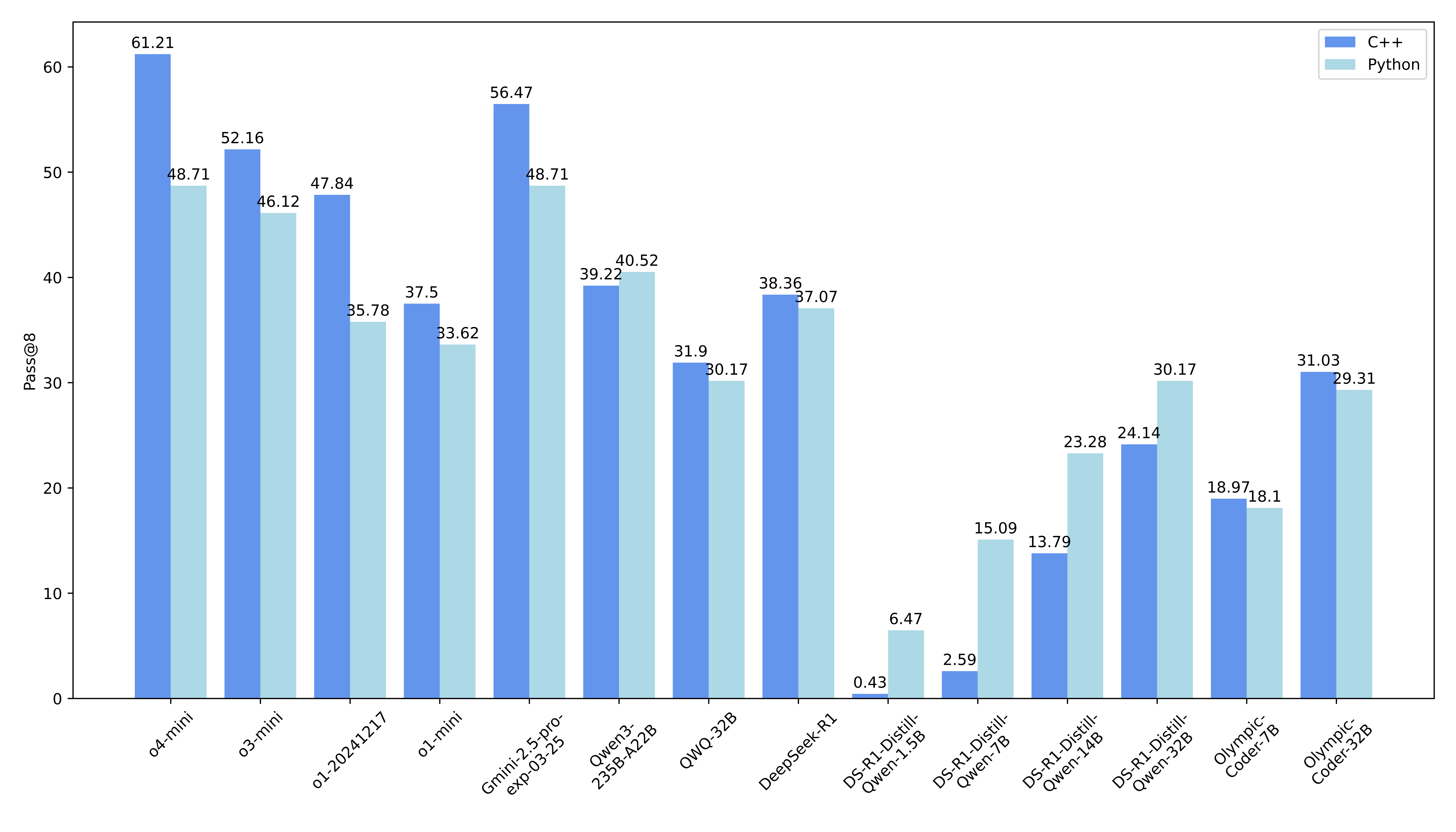}}
    \caption{The comparison between CPP and Python.}
    \label{fig:perf_on_py_cpp}
\end{figure*}

In addition to Python, we also evaluated the performance of the models using the CPP language on OJBench, the resluts are displayed in Figure \ref{fig:perf_on_py_cpp}.

The results indicate that for advanced reasoning-oriented models, such as o4-mini, o1-20241217, and Gemini-2.5-pro-exp-03-25, using CPP as the programming language yields significantly better performance on OJBench compared to Python. We assume that this is due to the fact that CPP is inherently a high-performance programming language, making it more suitable for solving competition-level programming tasks than Python.

For the Qwen series models distilled from DeepSeek-R1, the performance of these models using Python is much higher than that of CPP, while the performance of Olympic-Coder-7B/32B using CPP and Python is not much different, with CPP slightly higher than Python. We attribute this to differences in the training data. The former models were trained using a large amount of distilled data from DeepSeek-R1, while the Olympic-Coder-7B/32B models were trained using CPP solutions for competition-level problems from the same teacher model, enabling them to better leverage CPP to solve tasks in OJBench. We provide more details on the pass rates of  models using CPP and Python in Appendix \ref{appendix:more_details_for_cpp_vs_py}.

\subsection{Refinement can improve the performance of the model}
\begin{figure*}[h]
    \centering
    \resizebox{\linewidth}{!}{
    \includegraphics{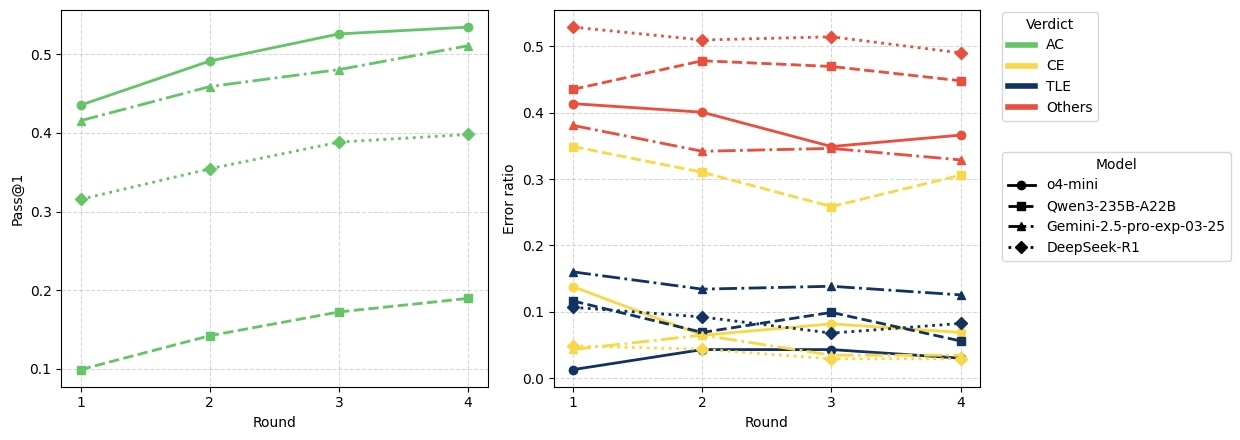}}
    \vspace{-0.3cm}
    \caption{Refinement on OJBench. Accepted (AC) indicates the model’s solution is correct. All error types include Wrong Answer (WA), Compile Error (CE), Memory Limit Exceeded (MLE), Time Limit Exceeded (TLE), RTE (Runtime Error), etc. Among all types, TLE occurs most frequently and CE  has no direct correlation with the reasoning ability of the model. Therefore, we distinguish \textbf{CE} and \textbf{TLE} from other error types  \textbf{ (Others)}. This allows us to more clearly understand which error types the model can reduce through refinement.}
    \label{fig:refine}
\end{figure*}
In real-world competitive programming scenarios, human participants have access to error messages generated during code execution, which they use to debug and correct their code. Motivated by this, we explored whether models could leverage error messages from code execution to rectify erroneous solutions. Specifically, for erroneous model solutions, we directly utilize the solution code and its error feedback (e.g., Compile Error (CE), Time Limit Exceeded (TLE), Wrong Answer (WA)) as prompts to guide subsequent refinement. The results are shown in Figure \ref{fig:refine}.

We observed that through each round of refinement, the model can continuously improve the pass rate of the solution. Among different types of errors, the proportion of CE errors decreased most significantly. We assume that this is because CE are not directly related to the model's ability to solve competitive programming tasks but rather stem from the model's oversight of certain details during the coding process. Therefore, they can be easily corrected using the error messages.

However, we found that the model struggled to address TLE errors during the refinement process. This is because resolving TLE requires the model to design more efficient algorithms and data structures tailored to specific programming tasks, which is essential for solving complex competitive programming problems. This shows that models still face significant challenges in designing efficient algorithms to solve complex code reasoning tasks.

\subsection{Analysis of problems that the model cannot solve}
\begin{figure*}[h]
    \centering
    \resizebox{\linewidth}{!}{
    \includegraphics{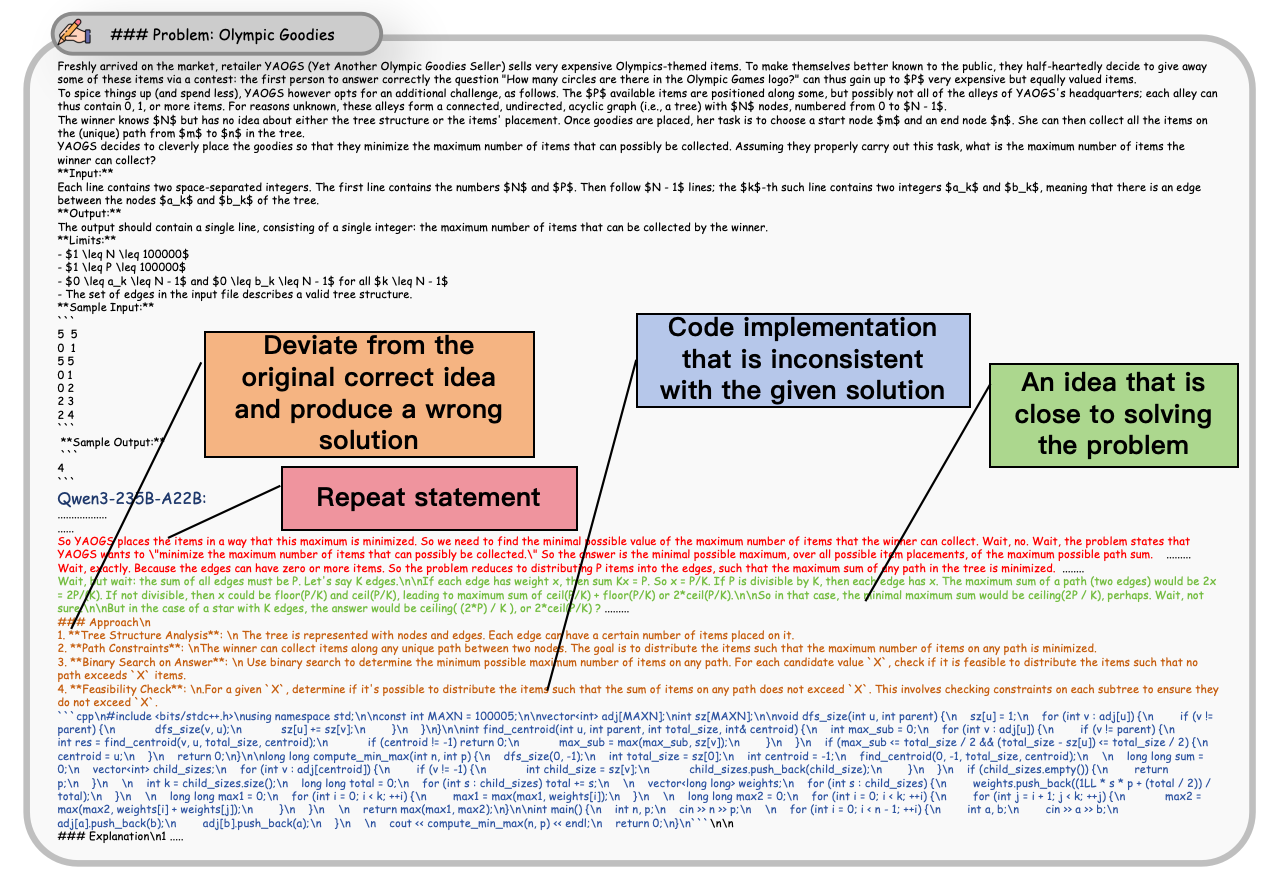}}
    \vspace{-0.3cm}
    \caption{Case analysis on OJBench}
    \vspace{-20pt}
    \label{fig:case1}
\end{figure*}
To thoroughly investigate the problem-solving strategies of large language models with reasoning capabilities, we selected Qwen3-235B-A22B for case analysis. This selection was made due to the inaccessibility of the internal reasoning processes of closed-source models. Figure \ref{fig:case1} presents a detailed case analysis of the reasoning process of Qwen3-235B-A22B. 

In this case, the model exhibited a significant amount of repetitive restatements of the problem requirements during the reasoning process, rather than engaging in a deeper analysis. Subsequently, the model developed a solution approach that was close to the correct line of reasoning. However, it failed to accurately assess the feasibility of this approach in subsequent steps, ultimately leading to an incorrect problem-solving strategy. Moreover, the code implemented by the model did not align with the proposed solution approach.

\section{Related work}
\textbf{Code Large Language Models. } In recent years, numerous large language models specifically designed for code-related tasks have emerged, such as AlphaCode\cite{li2022competition}, StarCoder\cite{li2023starcoder}, Magicoder\cite{wei2023magicoder}, and Qwen2.5-Coder\cite{hui2024qwen2}. These models, trained on extensive code corpora, have demonstrated remarkable capabilities in tasks related to code generation\cite{yu2024codereval}, completion\cite{chai2024mceval}, and debugging\cite{liu2024mdeval}. However, while these code LLMs have achieved significant improvements in simple code tasks, such as code completion and repair, they exhibit limited reasoning abilities when dealing with complex programming problems, such as those encountered in competitive coding scenarios.

\textbf{Reasoning-oriented Large Language Models. } Recently reasoning-oriented Large Language Models have demonstrated formidable reasoning capabilities across various domains \cite{team2025kimi,liu2025fin,diao2025driverx}, particularly those trained with large-scale reinforcement learning , such as OpenAI-o1\cite{jaech2024openai}, o3\cite{Day12ofS75:online}, DeepSeek-R1\cite{guo2025deepseek}, and QWQ\cite{qwq32b}. These models have shown reasoning abilities comparable to those of human competitive programmers in complex tasks involving mathematics and coding, rendering existing evaluation benchmarks insufficient for accurately assessing model performance. This makes it more urgent to develop competition-level evaluation datasets.

\textbf{Code Generation Benchmarks. } Recently, a growing number of benchmark datasets have been introduced to comprehensively evaluate the capabilities of language models \cite{qiao2024we,song2024cs,diao2025seas}. Among them, code generation benchmarks, such as HumanEval\cite{chen2021codex}, MBPP\cite{austin2021program}, and DS-1000\cite{lai2023ds}, have primarily focused on evaluating models' abilities to generate simple functions, perform data operations, and conduct mathematical computations. However, these benchmarks fail to provide a discriminative assessment for advanced Reasoning Large Language Models. Additionally, there are benchmarks specifically designed to evaluate models' performance on competition-level code generation problems, such as TACO\cite{li2023taco}, CodeELO\cite{quan2025codeelo}, Probench\cite{yang2025probench}, and USACO\cite{shi2024can}. These benchmarks are more complex than their predecessors and place higher demands on models' reasoning abilities. Nevertheless, these competitive benchmarks mainly focus on problems from coding platforms like CodeForces and LeetCode, without incorporating top-tier human algorithm competition problems. Our proposed OJBench focuses on the premier human programming competitions, NOI and ICPC, aiming to truly explore the limits of current large language models' code reasoning capabilities. We further categorize the problems in OJBench into different difficulty levels and support bilingual evaluation to better diagnose model performance and provide guidance for future code LLM development.

\section{Conclusion}
In this work, we introduce OJBench, a competition-level code reasoning benchmark. Our evaluation of 37 models reveals that even state-of-the-art reasoning-oriented models face significant challenges when tackling complex code reasoning tasks. Furthermore, we conducted in-depth experiments and analyses on the models based on this dataset, with the aim of providing valuable insights for future code LLM development.

\clearpage

\bibliography{neurips_2025}
\bibliographystyle{unsrt}
\appendix
\section{Limitations and Broader Impact}
\label{appendix_limitation}
This paper introduces OJBench, which is used to evaluate the code reasoning abilities of LLMs and provides guidance for the research field. However, this paper still has some limitations.

\textbf{Limitations.}  (1) Limited coverage: The data of OJBench mainly comes from open-source programming competition platforms and official open-source data from competitions, and it is of high difficulty. Although our benchmark is relatively more difficult compared to other benchmarks, given the breadth of the code specialization field, our evaluation cannot cover the entire scope of programming competitions. (2) Insufficient diversity: The problem types of OJBench are mainly focused on the field of algorithm competitions. For some emerging programming application scenarios, such as the development combining artificial intelligence and code, Internet of Things programming, and blockchain smart contract development, there is insufficient support for evaluating code reasoning abilities.

\textbf{Broader Impact.}  In the research field, OJBench is expected to play an important role in the field of computer science. With the help of OJBench, researchers can accurately evaluate the performance of LLMs in human programming competitions, thereby promoting the development of scientific research.

\section{Ethical Statement}
\label{Ethical_Statement}
We ensure that we comply with applicable laws and ethical standards during the data collection and usage processes, and provide adequate compensation to all crowd workers. Since this benchmark involves objective knowledge and reasoning in the field of programming, the annotation content is not affected by geographical or cultural differences among annotators.

Furthermore, our dataset does not contain any personally identifiable information or offensive content. The authenticity and accuracy of OJBench have been thoroughly verified, providing a reliable basis for evaluating LLMs. OJBench is intended solely for academic and research purposes. Any commercial use or other misuse that deviates from this purpose is strictly prohibited. We will urge all users to respect this regulation in order to maintain the integrity and ethical use of this valuable resource.

\section{Details of NOI and ICPC}
\label{appendix:Details of NOI and ICPC}

\textbf{NOI.}  The National Olympiad in Informatics is one of the five major Olympiads in China. It is characterized by its high level of difficulty, encompassing a wide range of knowledge areas such as algorithms, data structures, combinatorial mathematics, and computational geometry. The problems in NOI are highly flexible, focusing not only on the contestants' grasp of fundamental computer science knowledge but also on their logical thinking abilities in solving complex problems, as well as the depth of their thinking and the breadth of their knowledge.

\textbf{ICPC.}  The International Collegiate Programming Contest  is one of the most influential collegiate computer programming contests globally. This competition is conducted in teams, with each team consisting of three participants who are required to solve a series of complex programming problems within a specified time frame. The organizational structure of the ICPC includes regional contests and a global final.

\section{The pass rates of different difficulty levels of models using Python and CPP on OJBench}
\begin{figure*}[h]
    \centering
    \resizebox{\linewidth}{!}{
    \includegraphics{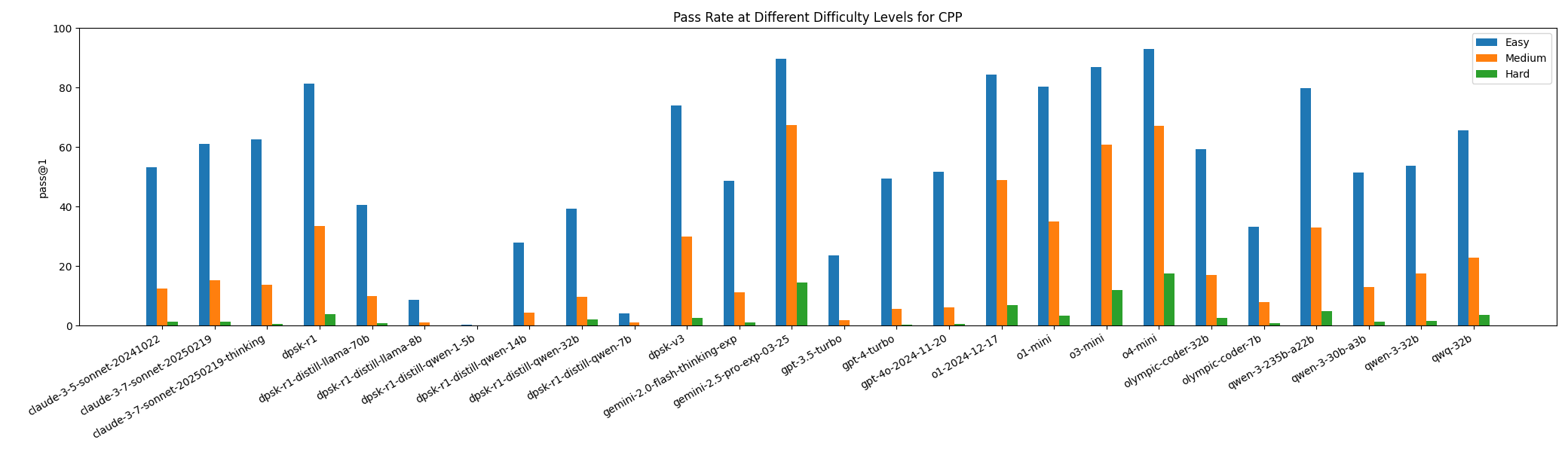}}
    \vspace{-0.3cm}
    \caption{Pass rate at different difficulty levels of more models using CPP on OJBench}
    \label{fig:perf_on_difficulty_cpp}
\end{figure*}

\begin{figure*}[h]
    \centering
    \resizebox{\linewidth}{!}{
    \includegraphics{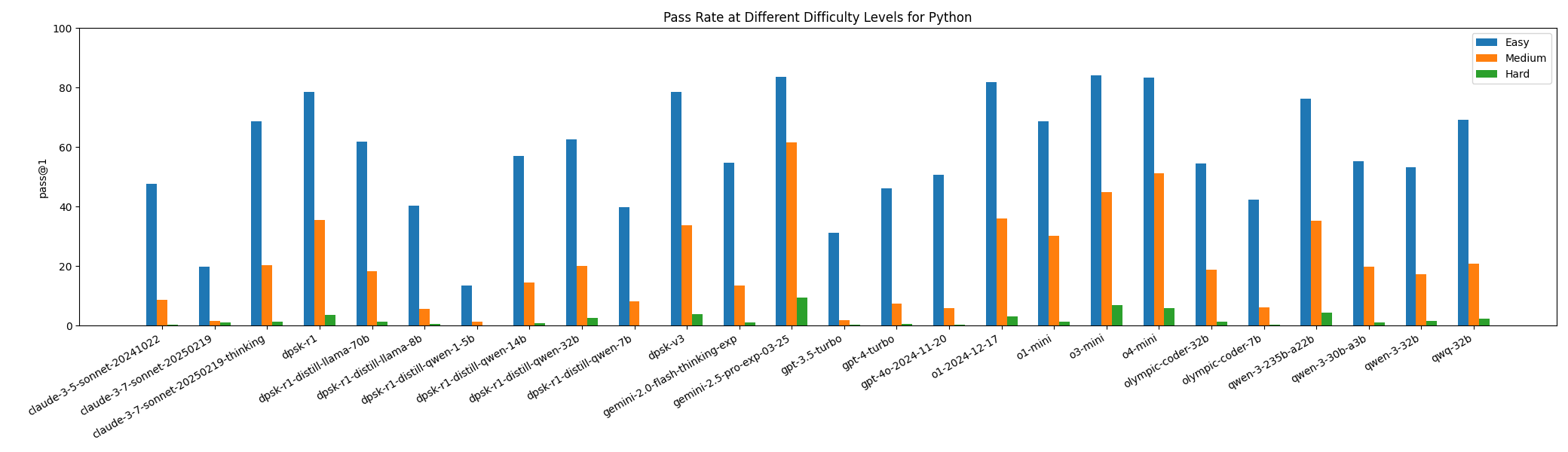}}
    \vspace{-0.3cm}
    \caption{Pass rate at different difficulty levels of more models using Python on OJBench}
    \label{fig:perf_on_difficulty_py}
\end{figure*}

Figure \ref{fig:perf_on_difficulty_cpp} and \ref{fig:perf_on_difficulty_py} show the pass rates of different difficulty levels in OJBench for more models using Python and CPP.
\label{appendix:more_details_for_cpp_vs_py}

\section{Details about the inference experiments on the open source models}
For inference experiments on all open source models with less than 72B parameters, we used two computing clusters equipped with 8 NVIDIA A100-80GB GPUs. Inference for each model took about two hours.


\end{document}